\documentclass{article} 
\usepackage{iclr2024_conference,times}

\usepackage{amsmath,amsfonts,bm}

\def\eqref#1{equation~\ref{#1}}

\def\1{\bm{1}}

\DeclareMathAlphabet{\mathsfit}{\encodingdefault}{\sfdefault}{m}{sl}
\SetMathAlphabet{\mathsfit}{bold}{\encodingdefault}{\sfdefault}{bx}{n}

\usepackage{hyperref}
\usepackage{url}
\usepackage{adjustbox}
\usepackage{lipsum}
\usepackage{color}
\usepackage{wrapfig}
\usepackage{booktabs}
\usepackage{multirow,mathtools} 
\usepackage{algorithm,algpseudocode}
\usepackage{threeparttable}
\usepackage{amsthm}
\usepackage{amsmath}
\usepackage{amssymb}
\usepackage{booktabs}
\usepackage{amsfonts}
\usepackage{wrapfig}
\usepackage{times}
\usepackage{multirow}
\usepackage{url}
\usepackage{color}
\usepackage{xcolor}
\usepackage{tabularx}
\usepackage{caption}

\usepackage{enumitem}
\usepackage{subcaption}
\usepackage{natbib}
\usepackage{wrapfig}
\usepackage{resizegather}
\usepackage{algorithm}
\usepackage{algpseudocode}
\usepackage{bbm}
\usepackage{makecell}
\usepackage{wrapfig}
\usepackage{mdframed}
\usepackage[many]{tcolorbox}
\newtcolorbox{questionblock}{
    enhanced,
    boxrule=0pt,frame hidden,
    interior style={
        top color=lightgray,
        bottom color=lightgray
    },
    colback=lightgray,
    colframe=white,
    sharpish corners,
    borderline north={0pt,none,white},
    borderline south={0pt,none,white},
    borderline east={0pt,none,white},
    borderline west={0pt,none,white},
}

\usepackage{pifont}

\title{Neuromorphic Computing with Multi-Frequency Oscillations: A Bio-Inspired Approach to Artificial Intelligence}

\author{Boheng Liu$^{1}$,
  Ziyu Li$^{1}$,
  Qing Li$^{1}$,
  Xia Wu$^{1}$ \\
  $^1$Beijing Institute of Technology \\
  \texttt{boheng@bit.edu.cn} \\
\\
}

\iclrfinalcopy 
\begin{document}

\maketitle

\begin{abstract}
Despite remarkable capabilities, artificial neural networks exhibit limited flexible, generalizable intelligence. 
This limitation stems from their fundamental divergence from biological cognition that overlooks both neural regions' functional specialization and the temporal dynamics critical for coordinating these specialized systems.
We propose a tripartite brain-inspired architecture comprising functionally specialized perceptual, auxiliary, and executive systems. Moreover, the integration of temporal dynamics through the simulation of multi-frequency neural oscillation and synaptic dynamic adaptation mechanisms enhances the architecture, thereby enabling more flexible and efficient artificial cognition.
Initial evaluations demonstrate superior performance compared to state-of-the-art temporal processing approaches, with 2.18\% accuracy improvements while reducing required computation iterations by 48.44\%, and achieving higher correlation with human confidence patterns.
Though currently demonstrated on visual processing tasks, this architecture establishes a theoretical foundation for brain-like intelligence across cognitive domains, potentially bridging the gap between artificial and biological intelligence.
\end{abstract}

  
\section{Introduction}
Artificial neural networks have achieved remarkable success, however, fundamental differences remain between their underlying architecture and that of biological cognition \cite{zhang2025inter}. 
These architectural limitations constrain the development of artificial intelligence in terms of flexibility, generalization ability, and interpretability key attributes essential for systems designed to engage in meaningful interactions with humans. 
Biological intelligence, characterized by its high flexibility and robustness \cite{oby2025dynamical}, demonstrates irreplaceable value in real-world applications and serves as a foundational reference for the advancement of neural networks. 
Drawing upon biological cognitive architectures is expected to bring artificial neural networks closer to human-level flexibility, adaptability, and general intelligence.

Biological cognition constitutes the core of biological intelligence. Having evolved over hundreds of millions of years, it demonstrates clear functional specialization and coordinated operational mechanisms within the brain  \cite{tiezzi2025back}. 
Neuroanatomical evidence reveals consistent organizational patterns across human brains, where distinct systems responsible for sensory processing, executive control, and modulatory functions work in concert to generate intelligent behaviors. 
In contrast, most existing artificial networks focus more on improving computational processes rather than adopting organizational perspectives that mimic biological principles \cite{schulze2025visual, wang2024system}. 
They lack a complex and integrated organizational structure similar to that of the brain, which may be a key reason for their performance limitations \cite{masset2025multi}. 
To bridge this gap, it is essential to reconceptualize neural architectures by incorporating principles derived from brain organization.

Furthermore, the coordinated operation of these functionally specialized brain systems critically relies on complex temporal mechanisms, including multi-frequency neural oscillations \cite{buzsaki2004neuronal, caglayan2023molecular} and synaptic dynamic adaptation\cite{shen2023brain}, which ensure their integrated functioning. 
Various artificial neural networks attempt to handle temporal dependencies, from recurrent architectures like LSTM \cite{shi2015convolutional} and Transformers \cite{vaswani2017attention} that capture long-range dependencies, to adaptive computation methods \cite{banino2021pondernet, xue2023adaptive} that adjust processing based on input complexity. 
More recently, the Continuous Thought Machine framework (CTM) \cite{darlow2025continuous} has made substantial progress by explicitly incorporating temporal processing and neural synchronization as core representational mechanisms. 
However, these approaches lack mechanisms that integrate specialized processing with temporal dynamics across multiple timescales, a fundamental limitation that restricts their ability to model the complex interactions essential to biological cognition.

\begin{figure}[t]
\centering
\includegraphics[width=\columnwidth]{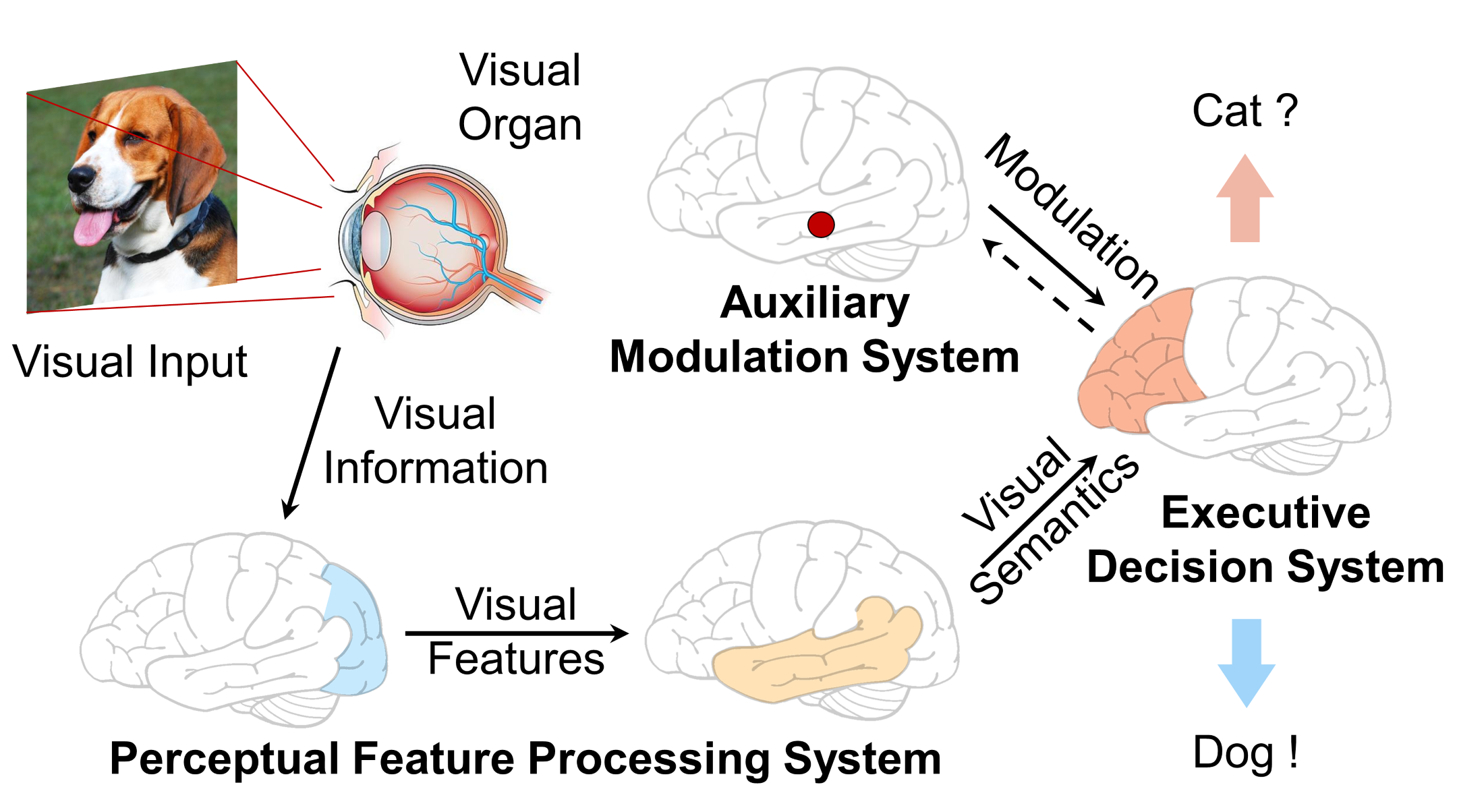} 
\caption{Tripartite Brain Cognitive Architecture in the Human Brain.
In visual tasks, three functionally specialized systems Perceptual (visual cortex and temporal lobe), Auxiliary (ventral tegmental area), and Executive (frontal lobe) reflect the regional collaboration of biological neural organization.
Visual semantics from perceptual feature processing, while synergistic interactions between frontal lobe synaptic adaptation and ventral tegmental neuromodulation collectively support categorical decision-making.}
\label{fig1}
\end{figure}

In light of the aforementioned challenges, we propose a novel Tripartite Brain-Inspired Architecture as a foundational theoretical framework for bridging artificial and biological cognition. 
Our architecture organizes neural computation into three functionally specialized but interacting systems: a Perceptual Feature Processing System analogous to sensory cortical regions, an Auxiliary Modulation System reflecting subcortical modulatory structures, and an Executive Decision System corresponding to prefrontal cortical areas. 
As illustrated in Figure \ref{fig1}, this tripartite organization represents the minimal paradigm necessary to capture fundamental biological brain organization principles and provides a comprehensive framework for understanding how different neural processes collaborate to produce intelligent behavior. 
By organizing computation according to these principles, our architecture establishes a theoretical foundation for addressing current limitations in artificial intelligence systems.

Building upon this foundational architecture, we introduce two complementary technical innovations that directly implement the critical temporal mechanisms found in biological brains. 
First, we develop a Synaptic Dynamic Adaptation mechanism that enhances the Executive Decision System by incorporating adaptive neural connectivity modulation based on input complexity, similar to biological spike timing dynamics in prefrontal cortical circuits.
Second, we develop a Neural Oscillation and Neuromodulation mechanism that enhances the Auxiliary Modulation System by incorporating multi-frequency temporal synchronization and context-sensitive parameter regulation, similar to biological neuromodulatory systems.
These innovations demonstrate how our tripartite architecture can effectively integrate temporal processing capabilities absent in conventional approaches.

Our empirical investigations on image classification tasks demonstrate that the Tripartite Brain Inspired Architecture achieves superior performance compared to state-of-the-art approaches including CTM, with accuracy improvements of up to 2.18\% and iteration reductions of up to 48.44\%, while maintaining robustness to noise-corrupted inputs and demonstrating stronger correlation with human categorization patterns.
By establishing a comprehensive architectural framework that incorporates both functional specialization and temporal dynamics, our work represents a significant advancement toward developing artificial systems that more closely reflect biological intelligence principles, potentially bridging the gap between artificial and biological cognition.

\section{Related Work}
\subsection{Cognitive theories of Functional Specialization}
Neuroscientific evidence reveals three functionally specialized systems that provide the foundation for our architecture. 
The perceptual system, comprising sensory cortices through association areas, processes information in increasingly abstract hierarchies \cite{mesulam1998sensation}. 
The auxiliary system, including subcortical structures such as the ventral tegmental area and amygdala, provides modulatory control through neuromodulatory mechanisms that adjust processing based on context \cite{bechara2000emotion}. 
The executive system in the prefrontal regions supports decision-making and cognitive control by integrating sensory inputs and flexibly adapting to environmental changes \cite{badre2018frontal}. 
This organization aligns with theories such as Global Workspace Theory, which proposes that consciousness emerges from information broadcasting among specialized modules \cite{baars1993cognitive, dehaene2003neuronal}. 
Although these biological principles have been extensively validated, most existing artificial neural networks implement these mechanisms in isolation, lacking a unified architectural design. 
This limitation prevents artificial neural networks from achieving the efficient information integration and flexible environmental adaptability observed in biological brains. 
Our architecture systematically transforms the tripartite model of \textit{perceptual processing, executive integration, and auxiliary modulation} into a cohesive framework across multiple levels of abstraction.

\subsection{Temporal Neural Processing Models}
Temporal dynamics integration has progressed through several distinct computational approaches. 
LSTM \cite{shi2015convolutional} introduced a gating mechanism for time series analysis, but with limited interpretability.
PonderNet \cite{banino2021pondernet} implemented adaptive computation by dynamically allocating resources based on input complexity. 
Liquid Time-Constant Networks \cite{hasani2021liquid} employed time-varying differential equations enabling neurons to adapt to input history. 
The CTM \cite{darlow2025continuous} advanced this field by implementing variable processing durations as explicit representational mechanisms. 
Despite these innovations, current models primarily focus on sequence processing rather than intrinsic neural dynamics, which limits their modeling efficiency and accuracy. 
Neuronal populations in the brain generate rhythmic oscillations through synchronized electrical activity \cite{caglayan2023molecular}. They achieve information binding across different time scales by synchronizing rhythms at various frequencies such as $\theta$ and $\gamma$ waves \cite{buzsaki2004neuronal}, while modulating synaptic plasticity to dynamically adjust integration precision \cite{shen2023brain}. 
Current artificial intelligence systems largely overlook these temporal coordination mechanisms. Our architecture incorporates synaptic dynamic adaptation, neural oscillation, and neuromodulation, enabling dynamic adjustment of complexity while maintaining temporal coherence across specialized processing systems.

\begin{figure*}[t] 
\centering
\includegraphics[width=\linewidth]{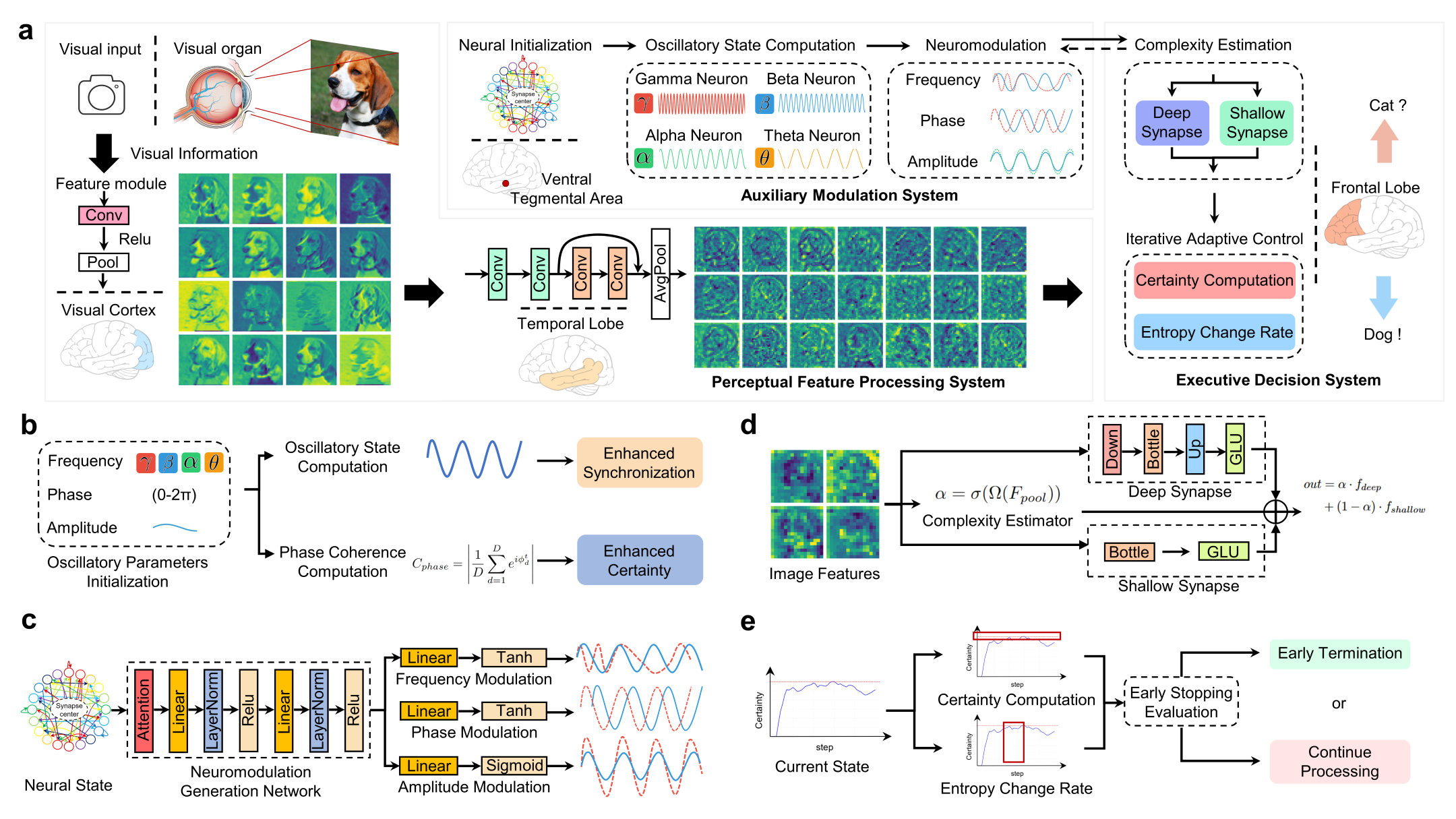} 
\caption{Tripartite Brain-Inspired Architecture with neural oscillation and adaptive processing.
\textbf{a}, Information flow from visual input through Perceptual (visual cortex and temporal lobe), Auxiliary (ventral tegmental area), and Executive (frontal lobe) systems, integrating hierarchical processing with neural dynamics for categorical output. 
\textbf{b}, Neural oscillation mechanisms with frequency band assignment ($\gamma $, $\beta $, $\alpha $, $\theta $).
\textbf{c}, Neuromodulation network that adjusts oscillatory parameters based on context. 
\textbf{d}, Adaptive processing showing complexity-based synaptic pathway selection.
\textbf{e}, Iterative control mechanism that dynamically determines computation termination based on certainty metrics.}
\label{fig2}
\end{figure*}
 
\section{Methods}

\subsection{Framework Overview}
Our Tripartite Brain-Inspired Architecture organizes neural computation into three functionally specialized systems that mirror fundamental organizational principles observed in biological brains. 
This comprehensive framework extends beyond temporal processing to establish a principled approach for integrating different aspects of neural computation. 
Figure 2a illustrates the complete architecture and information flow through these components.
The three primary systems, Perceptual Feature Processing System (PFPS), Auxiliary Modulation System (AMS), and Executive Decision System (EDS), fulfill distinct roles while maintaining continuous interaction, creating an integrated cognitive architecture analogous to the functional organization of human brains.

\subsection{Tripartite Brain-Inspired Architecture}

The Tripartite Brain-Inspired Architecture consists of three interconnected functional systems that process information in a coordinated manner. Here, we mathematically formalize their operations and interconnections.

\subsubsection{Perceptual Feature Processing System.}

The PFPS, analogous to sensory cortical regions, handles the initial feature extraction from sensory inputs ${x} \in {R}^{H \times W \times C}$. 
This system processes input through a backbone encoder $\Phi$ (typically a ResNet variant) and applies a positional embedding function $P$ to incorporate spatial information:

\begin{equation}
{F} = \Phi({x}) + P(\Phi({x}))
\end{equation}

\noindent
where ${F} \in {R}^{h \times w \times d}$ represents the extracted features. 
These features are subsequently flattened and projected to form key-value pairs:

\begin{equation}
{k}^t = {W}_k{F}_{{flat}} \in {R}^{N \times d_k}
\end{equation}

\begin{equation}
{v}^t = {W}_v{F}_{{flat}} \in {R}^{N \times d_v}
\end{equation}

\noindent
where ${F}_{{flat}} \in {R}^{N \times d}$ is the flattened feature map with $N = h \times w$, and ${W}_k$ and ${W}_v$ are learnable projection matrices.

\subsubsection{Executive Decision System.}

The EDS corresponds to prefrontal cortical regions, integrating processed information to generate decisions. 
EDS maintains an internal state $\mathbf{z}^t \in {R}^{D}$ at each discrete step $t \in \{1, 2, ..., T\}$, enabling the iterative refinement of representations, even for static inputs. 
$D$ is the dimensionality of the representational space.
At each tick, the EDS generates query vectors from its internal state:

\begin{equation}
{q}^t = {W}_q{z}^t \in {R}^{d_q}
\end{equation}

These queries interact with perceptual features from the PFPS through an attention mechanism:

\begin{equation}
{o}^t = {Attention}({q}^t, {k}^t, {v}^t) = {softmax}\left(\frac{{q}^t({k}^t)^T}{\sqrt{d_k}}\right){v}^t \in {R}^{d_v}
\end{equation}

The EDS updates its internal state by integrating attention outputs with the previous state:

\begin{equation}
{a}^t = f_{\theta}({concat}({z}^t, {o}^t)) \in {R}^{D}
\end{equation}

\begin{equation}
{z}^{t+1} = {z}^t + {a}^t
\end{equation}

\noindent
where $f_{\theta}$ is a neural network that processes the concatenated state and attention output.
The EDS also maintains a memory of recent activations:

\begin{equation}
{A}^t = [{a}^{t-M+1} \; {a}^{t-M+2} \; \cdots \; {a}^t] \in {R}^{D \times M}
\end{equation}

\noindent
where $M$ is the memory length.

\subsubsection{Auxiliary Modulation System.}

The AMS, analogous to subcortical structures, provides regulatory functions that optimize processing. This system generates modulatory signals based on both current perceptual features and the EDS internal state:

\begin{equation}
{M}^t = \Psi({F}, {z}^t) \in {R}^{d_m}
\end{equation}

\noindent
where $\Psi$ is the modulation generation network. 
These modulatory signals affect processing through:
(1) Parameter modulation in the EDS update function:

   \begin{equation}
   f_{\theta}({concat}({z}^t, {o}^t)) \rightarrow f_{\theta \odot g({M}^t)}({concat}({z}^t, {o}^t))
   \end{equation}

   \noindent
   where $g(\cdot)$ transforms modulatory signals to parameter scaling factors and $\odot$ represents element-wise multiplication.
(2) Attention modulation:

   \begin{equation}
   {o}^t = {Attention}({q}^t, {k}^t, {v}^t, {M}^t)
   \end{equation}

   \noindent
   where modulatory signals influence attention weights.

\subsubsection{Inter-system Communication and Synchronization.}

Communication between systems is facilitated through synchronization patterns that enable information integration:

\begin{equation}
    S^t_{ij} = \frac{({z}^t_i)^T \cdot {diag}({R}^t_{ij}) \cdot {z}^t_j}{\sqrt{\sum_{\tau=1}^t [{R}^t_{ij}]_\tau}}
\end{equation}

\noindent
where ${z}^t_i$ and ${z}^t_j$ represent the activations of neurons $i$ and $j$ at time $t$, and ${R}^t_{ij} \in {R}^t$ contains exponential decay factors that modulate temporal dependencies.
The final output at each internal tick is computed as:

\begin{equation}
{y}^t = {W}_{{out}} \cdot {S}^t_{{out}} \in {R}^C
\end{equation}

\noindent
where ${W}_{{out}} \in {R}^{C \times N_s}$ is a learned weight matrix, ${S}^t_{{out}} \in {R}^{N_s}$ represents output synchronization patterns sampled from $S^t_{ij}$, and $C$ is the number of output classes.
This integrated formulation establishes a complete processing pipeline where perceptual features flow into the executive system, modulated by auxiliary signals, with clear mathematical relationships between all components.

\subsection{Enhancements to Auxiliary Modulation System}

\subsubsection{Multi-Frequency Neural Oscillations.}

As illustrated in Figure 2b, we assign neurons to different frequency bands, mirroring biological neural oscillations:

\begin{equation}
\omega_d = \omega_{{base},d} + \delta\omega_d
\end{equation}

\noindent
where $\omega_{{base},d}$ is the base frequency assigned to neuron $d$ (drawn from $\gamma$, $\beta$, $\alpha$, or $\theta$ bands corresponding to 40-100Hz, 13-30Hz, 8-12Hz, and 4-7Hz respectively), and $\delta\omega_d$ is a learnable frequency offset.
The phase $\phi^t_d$ and amplitude $A^t_d$ of each neuron evolve over internal ticks:

\begin{equation}
\phi^t_d = \phi^0_d + 2\pi\omega_d t
\end{equation}

\begin{equation}
A^t_d = A^0_d
\end{equation}

\noindent
where $\phi^0_d$ and $A^0_d$ are learnable initial phase and amplitude parameters.
The oscillatory state of each neuron is then computed:

\begin{equation}
{z}^t_{{osc},d} = A^t_d \sin(\phi^t_d)
\end{equation}

This oscillatory state is integrated with the standard activation state:

\begin{equation}
{z}^t_{{combined}} = {z}^t + \lambda \cdot {z}^t_{{osc}}
\end{equation}

\noindent
where $\lambda$ is a scaling factor controlling the influence of oscillations. 
This combined state ${z}^t_{{combined}}$ subsequently serves as the effective internal state for the EDS.

\subsubsection{Neuromodulatory System.}

As illustrated in Figure 2c, the neuromodulatory system generates context-dependent signals that regulate neural parameters:

\begin{equation}
{M}^t = \{\delta{\omega}^t, \delta{\phi}^t, \delta{A}^t\} = \Gamma({o}^t)
\end{equation}

\noindent
where $\Gamma$ represents the modulation generation network that processes attention output ${o}^t$ to produce modulation signals.
These modulation signals adjust neural oscillation parameters:

\begin{equation}
\begin{gathered}
\omega^t_d = \omega_d(1 + \kappa_\omega \cdot \delta\omega^t_d) \\
\phi^t_d = \phi^t_d + \kappa_\phi \cdot \delta\phi^t_d \\
A^t_d = A^t_d(1 + \kappa_A \cdot \delta A^t_d)
\end{gathered}
\end{equation}

\noindent 
where $\kappa_\omega$, $\kappa_\phi$, and $\kappa_A$ are scaling constants that control the strength of modulation. 
Critically, these modulated parameters influence both the oscillatory dynamics and indirectly affect the complexity estimation network in the EDS through the combined state representation.

\subsubsection{Certainty Measure.}
Furthermore, neural oscillations quantify organized and non-random activity patterns in neural signals through certainty measure, and these patterns are closely related to cognitive functions. 
We formalize certainty in our model through two complementary measures: synchronization strength captures neuronal synergy, while phase coherence ensures information transmission fidelity. 
This approach quantifies reasoning certainty by measuring the stability of iterative neuronal activity:

\begin{equation}
C_{{phase}} = \left|\frac{1}{D}\sum_{d=1}^{D}e^{i\phi^t_d}\right|
\end{equation}

\begin{equation}
C_{{total}}^t = \beta \cdot C_{{entropy}}^t + (1-\beta) \cdot C_{{phase}}^t
\end{equation}

\noindent
where $C_{{entropy}}^t$ is the standard 1-normalized entropy measure, $C_{{phase}}$ ranges from 0 (no synchronization) to 1 (perfect synchronization), and $\beta$ is a weighting parameter. 

\subsection{Enhancements to Executive Decision System}

\subsubsection{Synaptic Dynamic Adaptation (SDA).}

The SDA mechanism dynamically modulates synaptic efficiency, strengthening connections for complex inputs requiring accuracy while weakening them for simpler inputs. 
This biological principle of on-demand adjustment enhances EDS performance through two parallel processing pathways with different complexities (Figure 2d):

\begin{equation}
f_{\theta_{{syn\_deep}}} = {SynapseUNET}(D, {depth}, {width}, {dropout})
\end{equation}

\begin{equation}
f_{\theta_{{syn\_shallow}}} = {MLP}(D, {dropout})
\end{equation}

\noindent
where $f_{\theta_{{syn\_deep}}}$ represents a complex U-NET structure with multiple skip connections, while $f_{\theta_{{syn\_shallow}}}$ is a simple single-layer MLP with nonlinearity and normalization.
The architecture computes a complexity coefficient $\alpha$ for each input using a dedicated estimation network that incorporates both perceptual features:

\begin{equation}
\alpha = \sigma(\Omega(F_{{pool}}))
\end{equation}

\noindent
where $F_{{pool}}$ represents pooled features, and $\Omega$ is the complexity estimation network.
This coefficient determines the mixture of deep and shallow processing:

\begin{align}
    \begin{gathered}
{a}^t = \alpha \cdot f_{\theta_{{syn\_deep}}}({concat}({z}^t_{{i}}, {o}^t))\\
 + (1-\alpha) \cdot f_{\theta_{{syn\_shallow}}}({concat}({z}^t_{{i}}, {o}^t))
    \end{gathered}
\end{align}

Note that this activation computation now utilizes the oscillation-influenced combined state ${z}^t_{{i}}$ rather than the basic state, creating a direct pathway for oscillatory dynamics to influence executive processing.

\subsubsection{Iterative Adaptive Control (IAC).}

As illustrated in Figure 2e, the IAC mechanism dynamically determines the appropriate number of internal ticks for each input using both entropy dynamics and the certainty measures derived from neural oscillations. 
For each input, the architecture tracks entropy change over consecutive ticks:

\begin{equation}
\Delta H^t = |H^t - H^{t-k}|/k
\end{equation}

When both the entropy change falls below a threshold and the combined certainty measure ($C_{{total}}^t$) , the architecture can terminate computation early:

\begin{equation}
{stop}^t = (\Delta H^t < \epsilon) \land (C_{{total}}^t > \tau)
\end{equation}

\noindent
where $H^t$ is the entropy at tick $t$, $k$ is the window size for measuring change, $\epsilon$ is the entropy change threshold, and $\tau$ is the certainty threshold.
To ensure minimum processing, we enforce a minimum number of ticks $T_{{min}}$:

\begin{equation}
{actual\_stop}^t = {stop}^t \land (t \geq T_{{min}})
\end{equation}
 
\subsection{Neural Synchronization as Representation}
When incorporating oscillatory dynamics, we enhance the inter-neuron synchronization patterns with phase synchronization:

\begin{equation}
S^t_{ij} = \frac{(\mathbf{z}_i^t)^T \cdot \text{diag}(\mathbf{R}^t_{ij}) \cdot \mathbf{z}_j^t}{\sqrt{\sum_{\tau=1}^t [\mathbf{R}^t_{ij}]_\tau}} \cdot \left|\frac{1}{t}\sum_{\tau=1}^{t}e^{i\Delta\phi_{ij}^\tau}\right|
\end{equation}

\noindent
where $\mathbf{z}_i^t$ and $\mathbf{z}_j^t$ denote the combined embeddings for neurons $i$ and $j$ at time $t$, $\mathbf{R}^t_{ij}$ contains exponential decay factors that modulate temporal dependencies, and $\Delta\phi_{ij}^\tau = \phi^\tau_i - \phi^\tau_j$ represents the phase difference between neurons using their modulated phase values from the AMS.
From this synchronization matrix, we sample $(i,j)$ pairs to create a key synchronization representations:
${S}^t_{{out}} \in {R}^{N_s}$ for outputs, directly influencing class predictions through:

\begin{equation}
{y}^t = {W}_{{out}} \cdot {S}^t_{{out}}
\end{equation}

\subsection{Loss Function and Optimization}

For each input, the architecture produces class distributions ${y}^t$ and certainty measures $C_{{total}}^t$ at each internal tick $t$. 
The loss at each time step is defined as:

\begin{equation}
L_t = {L}_{{CE}}(\mathbf{y}^t, {y}_{{true}})
\end{equation}

\noindent
where ${L}_{{CE}}$ is the cross-entropy loss between predicted and true class distributions.
We optimize performance across the internal thought dimension by aggregating loss dynamically:

\begin{equation}
L = \frac{L_{t_1} + L_{t_2}}{2}
\end{equation}

\noindent
where $t_1 = {argmin}_t(L_t)$ is the point of minimum loss, and $t_2 = {argmax}_t(C_{{total}}^t)$ is the point of maximum certainty as defined by our combined certainty measure.

Our integrated training approach with a unified loss function creates synergy across all three systems of the Tripartite Architecture, enabling them to develop specialized yet complementary functionalities that mirror how sensory, executive, and modulatory regions in biological neural systems co-evolve to support unified cognitive capabilities.

\begin{table*}[!t]
\begin{center}
\caption{Comparative Classification Performance. Acc: Classification Accuracy (\%). Params: Model Parameters (millions). Steps: Computational Average Iterations Required.}
\label{tab_1}
\resizebox{\textwidth}{!}{
\begin{tabular}{c|ccc|ccc|ccc}
\hline
{} & \multicolumn{3}{c|}{CIFAR10} & \multicolumn{3}{c|}{CIFAR100} & \multicolumn{3}{c}{SVHN}\\
 
{Method}& Acc ($\uparrow$) & Params ($\downarrow$) & Steps ($\downarrow$) & Acc ($\uparrow$) & Params ($\downarrow$) & Steps ($\downarrow$) & Acc ($\uparrow$) & Params ($\downarrow$) & Steps ($\downarrow$)  \\
\hline
FF & 85.41\scalebox{0.7}{$\pm$0.17} & \textbf{11.44} & - & 51.07\scalebox{0.7}{$\pm$0.57} & \textbf{11.44} & - & 95.23\scalebox{0.7}{$\pm$0.15} & \textbf{11.44} & - \\
LSTM & 84.89\scalebox{0.7}{$\pm$0.15} & 14.79 & 50.00\scalebox{0.7}{$\pm$0.00} & 6.92\scalebox{0.7}{$\pm$0.24} & 14.79 & 50.00\scalebox{0.7}{$\pm$0.00} & 95.30\scalebox{0.7}{$\pm$0.13} & 14.79 & 50.00\scalebox{0.7}{$\pm$0.00} \\
CTM & 86.11\scalebox{0.7}{$\pm$0.10} & 12.05 & 50.00\scalebox{0.7}{$\pm$0.00} & 50.70\scalebox{0.7}{$\pm$0.07} & 12.05 & 50.00\scalebox{0.7}{$\pm$0.00} & 95.16\scalebox{0.7}{$\pm$0.03} & 12.05 & 50.00\scalebox{0.7}{$\pm$0.00} \\
Ours & \textbf{87.06\scalebox{0.7}{$\pm$0.12}} & 12.90 & \textbf{25.78\scalebox{0.7}{$\pm$0.25}} & \textbf{52.88\scalebox{0.7}{$\pm$0.14}} & 12.90 & \textbf{32.85\scalebox{0.7}{$\pm$0.47}} & \textbf{95.39\scalebox{0.7}{$\pm$0.08}} & 12.90 & \textbf{35.42\scalebox{0.7}{$\pm$0.53}}\\
\hline

\end{tabular}
}
\end{center}
\end{table*}

\begin{figure*}[t] 
\centering
\includegraphics[width=\linewidth]{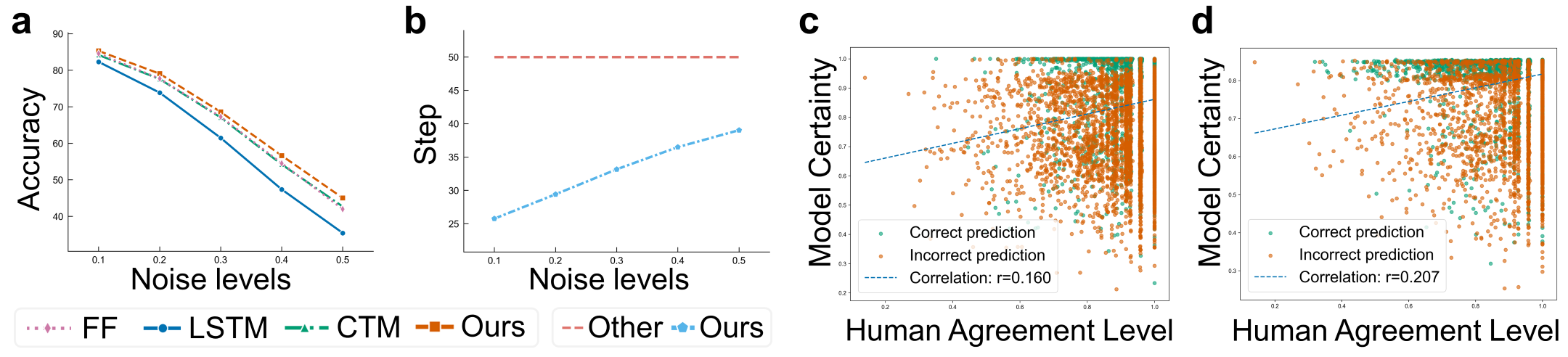} 
\caption{Robustness to input noise and human-alignment analysis. 
\textbf{a-b}, Accuracy and average iteration steps under different fixed gaussian noise levels ($\sigma$) for CIFAR10. \textbf{c}, Comparison between CTM and human categorization on CIFAR-10H. \textbf{d}, Comparison between our model and human categorization on CIFAR-10H.}
\label{fig3}
\end{figure*}

\section{Experiments} 

\subsection{Experimental Setup} 
\subsubsection{Datasets and Preprocessing.}

We selected four standard benchmark datasets with varying complexity characteristics: CIFAR-10, CIFAR-100, SVHN (Street View House Numbers), and CIFAR10-H (originally used to quantify human uncertainty). 
For CIFAR-10 \cite{krizhevsky2009learning}, CIFAR-100, and SVHN \cite{netzer2011reading}, we adhered to standard training and testing splits, with all images normalized to the range [0,1]. 
For CIFAR-10H \cite{peterson2019human}, we utilized human perceptual uncertainty labels for the 10,000 CIFAR-10 testset, containing normalized human classification probabilities across 10 categories from approximately 50 annotators per image, to evaluate our model's alignment with human categorization patterns.

\subsubsection{Implementation Details.}

All models were trained on NVIDIA L40s GPUs. For perceptual feature extraction, we employed a modified ResNet-18 \cite{he2016deep} backbone consistently across all models to ensure fair comparison. 
All models were trained for 100 epochs using the Adam optimizer with an initial learning rate of 1e-3 and cosine annealing schedule. We employed a batch size of 1024 across all experiments. 
To ensure robust evaluation, we conducted three independent runs with different random seeds.

\subsubsection{Baseline Methods.}
We compared our architecture against three representative approaches:
FF (Feed-Forward) \cite{vaswani2017attention}: Standard Transformer feed-forward architecture, representing traditional non-recurrent approaches;
LSTM \cite{shi2015convolutional} representing conventional recurrent architectures;
CTM \cite{darlow2025continuous} representing state-of-the-art adaptive computation approaches.

\subsection{Performance on Standard Benchmarks}

\subsubsection{Classification Performance.}
As shown in Table 1, while achieving accuracy improvements of 0.95-2.18\%, our approach reduces computational iterations by 37.32\% compared to CTM. 
These results confirm that organizing neural computation into specialized, temporally-coordinated systems reflecting biological brain principles creates substantial advantages in both performance and efficiency, advancing neural architectures toward more brain-like computational capabilities.

\subsubsection{Robustness and Adaptive Processing Under Uncertainty.}

Figure 3a illustrates the comparative performance of our Tripartite Architecture maintains a consistent performance advantage when subjected to increasing levels of Gaussian noise ($\sigma $ = 0.1-0.5). 
Figure 3b reveals our architecture's biologically-inspired adaptive processing. The system dynamically adjusts computational depth from 25.78 iterations for clean images to 39.04 iterations for highly corrupted inputs ($\sigma $ = 0.5). This emergent relationship between input complexity and processing depth mirrors neurobiological findings where neural activity persists longer when processing ambiguous stimuli \cite{wang2025foundation}.

\subsubsection{Alignment with Human Categorization Patterns.}

Figures 3c and 3d compare model certainty against human agreement levels on CIFAR-10H. Our Tripartite Architecture demonstrates stronger correlation with human judgment patterns (r=0.207) compared to CTM (r=0.160). This enhanced alignment suggests our brain-inspired temporal dynamics and functional specialization better capture the cognitive mechanisms underlying human visual categorization under uncertainty.

\begin{figure*}[t] 
\centering
\includegraphics[width=\linewidth]{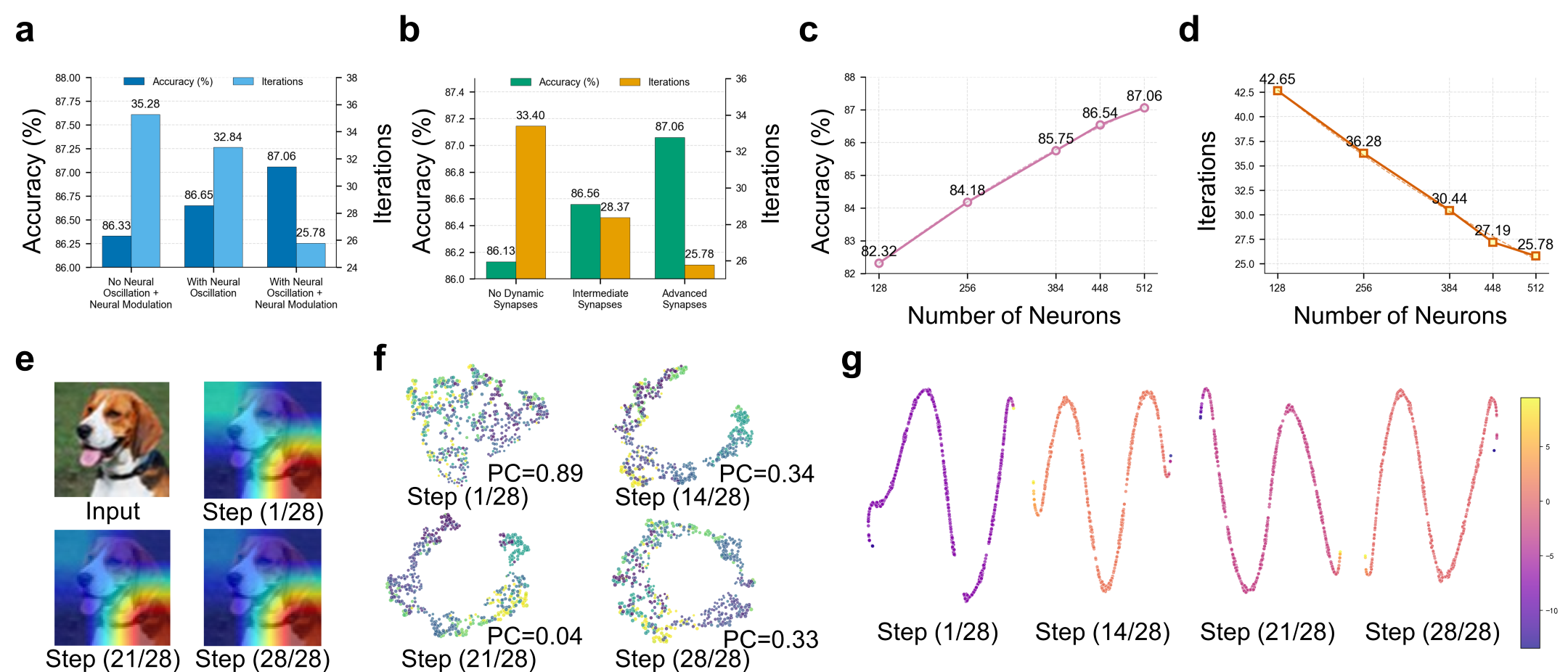} 
\caption{Ablation studies and visualization of Tripartite Brain-Inspired Architecture.
\textbf{a}, Impact of neural oscillation and modulation on performance and efficiency.
\textbf{b}, Effect of synaptic complexity on accuracy and computational iterations.
\textbf{c}, Accuracy improvement with increasing neuron density.
\textbf{d}, Computational efficiency gains with increasing neuron count.
\textbf{e}, Attention map visualization across early, middle, and final processing stages.
\textbf{f}, Phase coherence patterns showing characteristic changes throughout computation.
\textbf{g}, Evolution of neural activation patterns during iterative processing.}
\label{fig4}
\end{figure*}
  
\subsection{Ablation Studies}
\subsubsection{Neural Oscillation and Modulation Analysis.}

We evaluated three configurations of increasing oscillatory complexity as shown in Figure 4a. 
Each progressive addition from baseline to neural oscillations to full neuromodulation improved both accuracy (86.33\% to 87.06\%) and computational efficiency (35.28 to 25.78 iterations). 
Neuromodulation disproportionately enhanced processing efficiency relative to accuracy gains, mirroring biological systems where neuromodulatory pathways primarily regulate resource allocation rather than information content \cite{shine2021computational}.

\subsubsection{Synaptic Adaptation Analysis.} 
We investigated synaptic complexity effects as shown in Figure 4b.  
Progressive enhancement from basic to advanced configurations improved accuracy (86.13\% to 87.06\%) while reducing computational iterations (33.40 to 25.78).  
This positive correlation between synaptic complexity and performance parallels human neural systems \cite{insanally2024contributions, dellaferrera2022introducing}.

\subsubsection{Neuron Density Analysis.}
We examined how neuron count affects model performance as shown in Figures 4c and 4d. 
Increasing neuron density from 128 to 512 substantially improved accuracy (82.32\% to 87.06\%) while simultaneously reducing required computational iterations (42.65 to 25.78). 
This inverse relationship between neural resources and processing time parallels observations in biological systems, where increased neural allocation enables more efficient information processing through distributed computation \cite{kafashan2021scaling}.

\subsection{Visualization and Analysis}
 
Visualization of our architecture's internal dynamics reveals key operational mechanisms. 
Attention maps (Figure 4e) demonstrate progressive refinement from broad feature sampling to object-specific focus. 
Phase coherence (PC) analysis (Figure 4f) shows characteristic temporal signatures: high initial synchronization, middle-phase desynchronization during integration, and moderate resynchronization at decision convergence. 
Neural activation patterns (Figure 4g) exhibit cyclical behavior, with initially weak activations strengthening during intermediate processing before rhythmically alternating and finally stabilizing at completion. 
These visualizations reveal striking parallels to human neural oscillations, where synchronization dynamics in cortical regions coordinate information flow between specialized brain areas during perception and decision-making \cite{buzsaki2004neuronal}.
 
\section{Conclusion}
Our Tripartite Brain-Inspired Architecture bridges artificial neural networks and biological cognition through functionally specialized systems and temporal mechanisms, enhancing flexibility and generalizability. 
Empirical evaluations demonstrate significant advantages with 2.15\% accuracy improvements while reducing required computation iterations by 48.44\%, alongside superior robustness to noise and stronger correlation with human confidence patterns. 
Though validated on visual tasks, this framework provides a theoretical foundation for bridging artificial and biological intelligence across diverse cognitive domains including language processing, reasoning, and decision-making.

\bibliography{iclr2024_conference}
\bibliographystyle{iclr2024_conference}
\end{document}